\documentclass[conference]{IEEEtran}
\IEEEoverridecommandlockouts
\usepackage[show]{inotes}
\usepackage{cite}
\usepackage{amsmath,amssymb,amsfonts}
\usepackage{algorithmic}
\usepackage{graphicx}
\usepackage{textcomp}
\usepackage{xcolor}
\usepackage{multirow}
\usepackage{url}

\usepackage{booktabs}
\usepackage{breakurl}
\usepackage{listings}
\usepackage{multicol}
\usepackage{subfig}
\usepackage{graphics}
\usepackage{hyperref}
\usepackage{xstring}
\usepackage{graphicx}
\usepackage{tabularx}
\usepackage{pbox}
\usepackage{multirow}
\usepackage{enumerate}

\usepackage{amssymb}
\usepackage{pifont}

\def\BibTeX{{\rm B\kern-.05em{\sc i\kern-.025em b}\kern-.08em
    T\kern-.1667em\lower.7ex\hbox{E}\kern-.125emX}}

\usepackage{caption}
\usepackage{fancyhdr}
\usepackage{kantlipsum}
\fancypagestyle{plain}{
\fancyhf{}
\fancyhead[C]{Conference on \LaTeX} 
} \usepackage{eso-pic}

\begin{document}

\AddToShipoutPictureBG*{
\AtPageUpperLeft{
\setlength\unitlength{1in}
\hspace*{\dimexpr0.5\paperwidth\relax}
\makebox(0,-0.75)[c]{\textbf{2020 IEEE/ACM International Conference on Advances in Social Networks Analysis and Mining (ASONAM)}}}}


\title{
Deep Learning Benchmarks and Datasets for Social Media Image Classification for Disaster Response

\thanks{}
}



\author{
\IEEEauthorblockN{Firoj Alam\textsuperscript{1}, Ferda Ofli\textsuperscript{1}, Muhammad Imran\textsuperscript{1}, Tanvirul Alam\textsuperscript{2}, Umair Qazi\textsuperscript{1}}
\IEEEauthorblockA{\textsuperscript{1}Qatar Computing Research Institute, HBKU, Doha, Qatar\\
\textsuperscript{2}BJIT Limited, Dhaka, Bangladesh\\
\textsuperscript{1}\{fialam, fofli, mimran, uqazi\}@hbku.edu.qa, \textsuperscript{2}tanvirul.alam@bjitgroup.com}\\

}

\maketitle

\IEEEoverridecommandlockouts
\IEEEpubid{\parbox{\columnwidth}{\vspace{8pt}
\makebox[\columnwidth][t]{IEEE/ACM ASONAM 2020, December 7-10, 2020} \makebox[\columnwidth][t]{978-1-7281-1056-1/20/\$31.00~\copyright\space2020 IEEE} \hfill} \hspace{\columnsep}\makebox[\columnwidth]{}}
\IEEEpubidadjcol

\begin{abstract}
During a disaster event, images shared on social media helps crisis managers gain situational awareness and assess incurred damages, among other response tasks.
Recent advances in 
computer vision and deep neural networks have enabled 
the development of models for real-time image classification for a number of tasks, including 
detecting crisis incidents, filtering irrelevant images, classifying images into specific humanitarian categories, and assessing the severity of damage. 
Despite several efforts, past works mainly suffer from limited resources (i.e., labeled images) available to train more robust deep learning models.
In this study, we propose new datasets for disaster type detection, and informativeness classification, and damage severity assessment. Moreover, we relabel existing publicly available datasets for new tasks. We identify exact- and near-duplicates to form non-overlapping data splits, and finally consolidate them to create larger datasets. In our extensive experiments, we benchmark several state-of-the-art deep learning models and achieve promising results. We release our datasets and models publicly, aiming to provide proper baselines as well as to spur further research in the crisis informatics community.
\end{abstract}


\section{Introduction}
\label{sec:introduction}
Social media is widely used during natural or human-induced disasters as a source to quickly disseminate information and learn useful insights. People post content (i.e., through different modalities such as text, image, and video) on social media to get help and support, identify urgent needs, or share their personal feelings. Such information is useful for humanitarian organizations to plan and launch relief operations. As the volume and velocity of the content are significantly high, it is crucial to have real-time systems to automatically process social media content to facilitate rapid response.

There has been a surge of research works in this domain in the past couple of years. The focus has been to analyze the usefulness of social media data and develop computational models using different modalities to extract actionable information. Among different modalities (e.g., text and image), more focus has been given to 
textual content analysis 
compared to imagery content (see \cite{imran2015processing,said2019natural} for a comprehensive survey). Though many past research works have demonstrated that images shared on social media during a disaster event can help humanitarian organizations in a number of ways. For example, \cite{nguyen17damage} uses images shared on Twitter to assess the severity of the infrastructure damage and \cite{Mouzannar2018} focuses on identifying damages in infrastructure as well as environmental elements.

Current publicly available datasets for developing classification models for disaster response tasks include damage severity assessment dataset~\cite{nguyen17damage}, CrisisMMD~\cite{alam2018crisismmd}, and multimodal damage identification dataset~\cite{Mouzannar2018}. The annotated labels in these datasets include different classification tasks such as {\em (i)} disaster types, {\em (ii)} informativeness, {\em (iii)} humanitarian, and {\em (iv)} damage severity assessment. 
Upon studying these datasets, we note several limitations: {\em (i)} when considered independently, these datasets are fairly small in contrast to the datasets used in the computer vision community, e.g., ImageNet~\cite{deng2009imagenet} and MS COCO~\cite{coco_2014},
which entangles development of robust models for real-world applications, {\em (ii)} they contain exact- and near-duplicates, which often provides misleading performance scores due to the random train and test splits, {\em (iii)} inconsistent train/test splits have been used across different studies, which makes it difficult to compare the reported results in the literature. 
Another interesting aspect is that there has been significant progress in neural network architectures for image processing in the past few years; however, they have not been widely explored in the \textit{crisis informatics}\footnote{\url{https://en.wikipedia.org/wiki/Disaster_informatics}} for disaster response tasks. 

To address such limitations, our contributions in this study are as follows:
\begin{itemize}
    \itemsep0em
    \item We developed \emph{disaster types} and \emph{informativeness} datasets, which are completely new for the research community.
    \item We relabeled existing datasets for the new tasks. The motivation of using existing datasets for new tasks is that it significantly reduces the data collection, cleaning, and annotation efforts.
    Upon consolidation, we report the largest datasets available to date for different tasks.
    \item We divided each dataset into train, dev and test splits and created a \textit{non-overlapping test set}
    by eliminating exact- and near-duplicate images between the test and train sets. We also unified all task-specific datasets from different sources into a single set for different tasks.
    \item We provide benchmark results for four tasks, on separate as well as combined datasets, using several state-of-the-art neural network architectures. These results set new baselines for the crisis informatics community for the image classification tasks. 
    Finally, we will make the datasets with their splits publicly available.\footnote{https://crisisnlp.qcri.org/crisis-image-datasets-asonam20} 
\end{itemize}

The rest of the paper is organized as follows. Section~\ref{sec:related_works} provides a brief overview of the existing work. Section~\ref{sec:tasks} introduces the tasks while Section~\ref{sec:dataset} describes the datasets prepared for this study. Section~\ref{sec:experiments} explains the experiments and Section~\ref{sec:results_discussion_future_works} presents the results and discussion. Finally, we conclude the paper in Section~\ref{sec:conclutions}.

\section{Related Work}
\label{sec:related_works}


The studies on image processing in the crisis informatics domain are relatively fewer compared to the studies on analyzing textual content for humanitarian aid.\footnote{\url{https://en.wikipedia.org/wiki/Humanitarian_aid}} With recent successes of deep learning for image classification, research works have started to use social media images for humanitarian aid. The importance of imagery content on social media for disaster response tasks has been reported in many studies~\cite{petersinvestigating,daly2016mining,chen2013understanding,nguyen2017automatic,nguyen17damage,alam17demo,alam2019SocialMedia}. 
For instance, the analysis of flood images has been studied in \cite{petersinvestigating}, in which the authors reported that the existence of images with relevant textual content is more informative. Similarly, the study by Daly and Thom~\cite{daly2016mining} analyzed fire event's images, which are extracted from social media data. Their findings suggest that images with geotagged information are useful to locate the fire affected areas. 

The analysis of imagery content shared on social media has been recently explored using deep learning techniques for damage assessment purposes. Most of these studies categorize the severity of damage into discrete levels~\cite{nguyen2017automatic,nguyen17damage,alam17demo} whereas others quantify the damage severity as a continuous-valued index~\cite{nia2017building,li2018localizing}. 
Recently, \cite{alam2019SocialMedia} presented an image processing pipeline to extract meaningful information from social media images during a crisis situation, which has been developed using deep learning-based techniques. Their image processing pipeline includes collecting images, removing duplicates, filtering irrelevant images, and finally classifying them with damage severity. The study by Mouzannar et al.~\cite{Mouzannar2018} proposed a multimodal dataset, 
which has been developed for training a damage detection model. Similarly, \cite{multimodalbaseline2020} explores unimodal as well as different multimodal modeling approaches based on a collection of multimodal social media posts. 

Currently, publicly available datasets include damage severity assessment dataset~\cite{nguyen17damage}, CrisisMMD~\cite{alam2018crisismmd} and damage identification multimodal dataset~\cite{Mouzannar2018}. The former dataset is only annotated for images, whereas the latter two are annotated for both text and images.
Other relevant datasets are Disaster Image Retrieval from Social Media (DIRSM) \cite{bischke2017multimedia} and MediaEval 2018 \cite{alex258247}. 
For the image classification task, transfer learning has been a popular approach, where a pre-trained neural network is used to train a new model for a new task~\cite{yosinski2014transferable,sharif2014cnn,ozbulak2016transferable,oquab2014learning}. For this study, we follow same approach using different deep learning architectures. 

Our study differs from prior works in a number of ways. We propose new datasets for different tasks, annotate existing datasets for new tasks, create non-overlapping train/dev/test splits, and finally consolidate them to create a unified, large-scale dataset for several tasks. Lastly, we use the dataset to provide benchmarks using state-of-the-art deep learning models.

\section{Tasks}
\label{sec:tasks}
For this study, we addressed four different disaster-related tasks important for humanitarian aid. 

\subsection{Disaster type detection}
\label{ssec:task_disaster_types}
\begin{figure*}[htbp!]
	\renewcommand{\arraystretch}{0.6} 
	\linespread{0.5}\selectfont\centering
	\resizebox{0.8\linewidth}{!}{%
		\begin{tabular}{p{0.35\textwidth} p{0.35\textwidth} p{0.35\textwidth}}
			\includegraphics[width=0.33\textwidth]{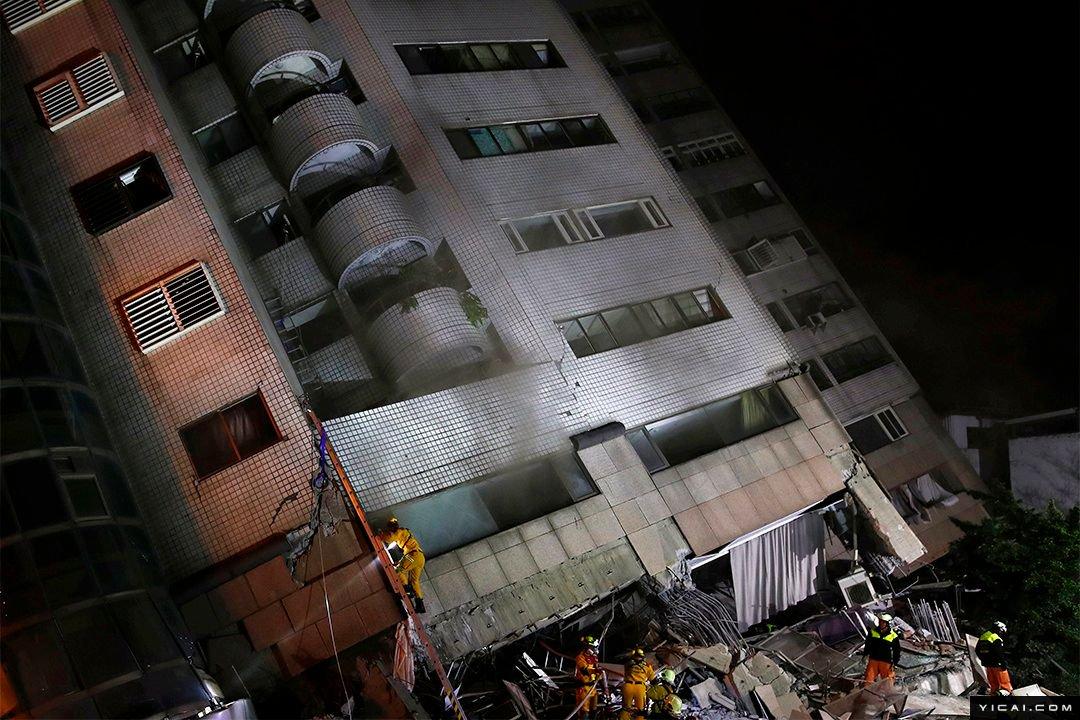}
			&
			\includegraphics[width=0.33\textwidth]{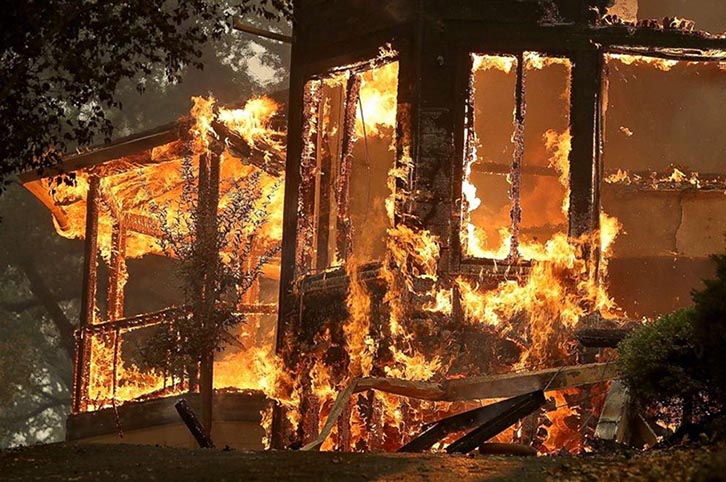}
			&
			\includegraphics[width=0.33\textwidth]{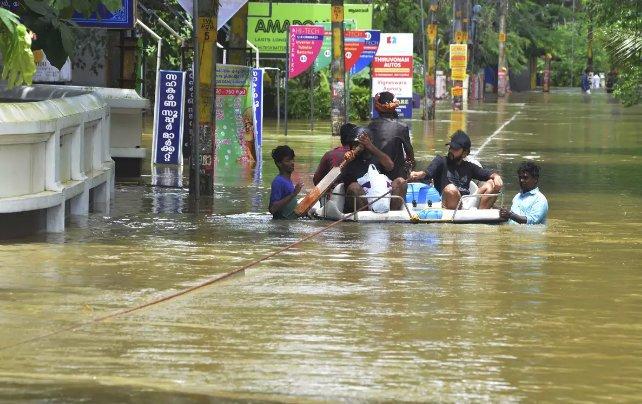}
			\\\\
			{\scriptsize \textbf{Earthquake}}
			&
            {\scriptsize \textbf{Fire}}
			&
            {\scriptsize \textbf{Flood}}
			\\\\
			\includegraphics[width=0.33\textwidth]{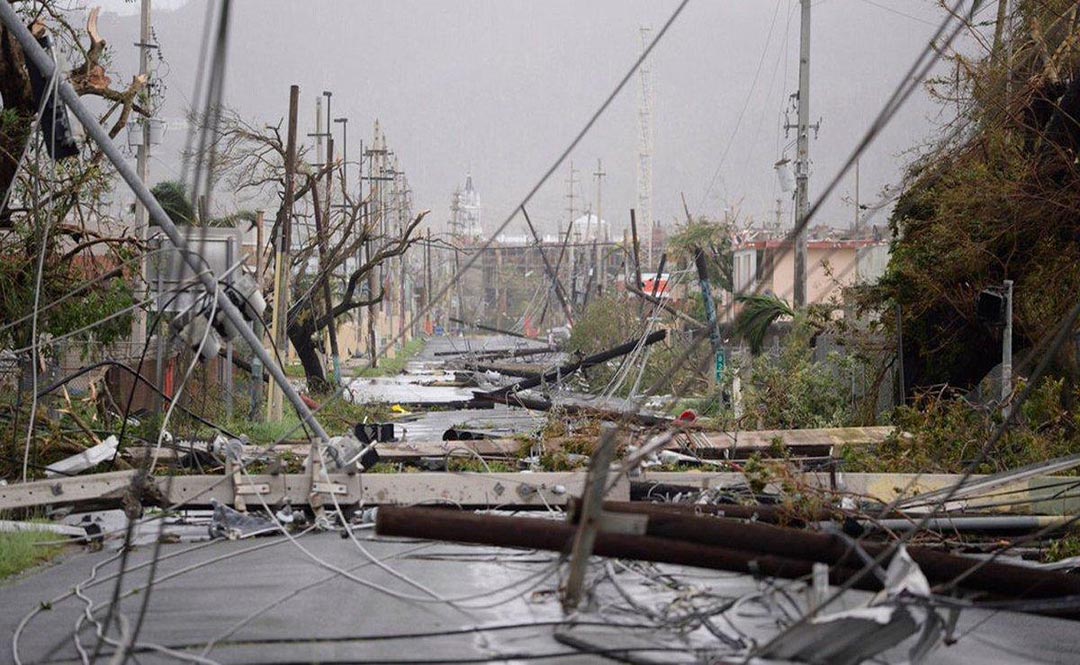}
			&
			\includegraphics[width=0.33\textwidth]{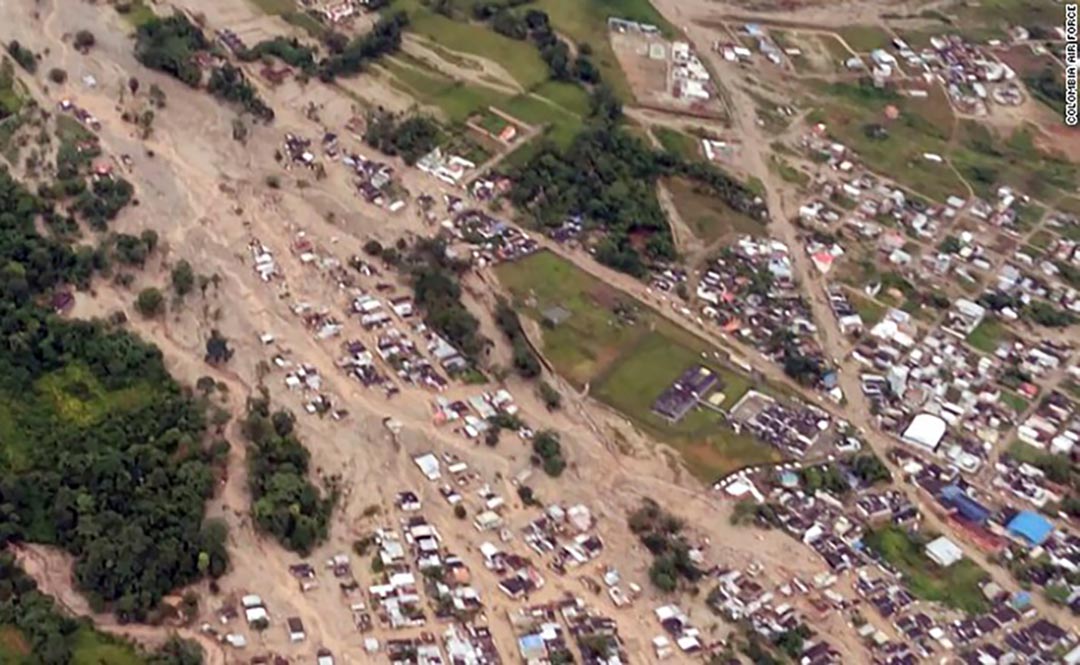}
			&
			\includegraphics[width=0.33\textwidth]{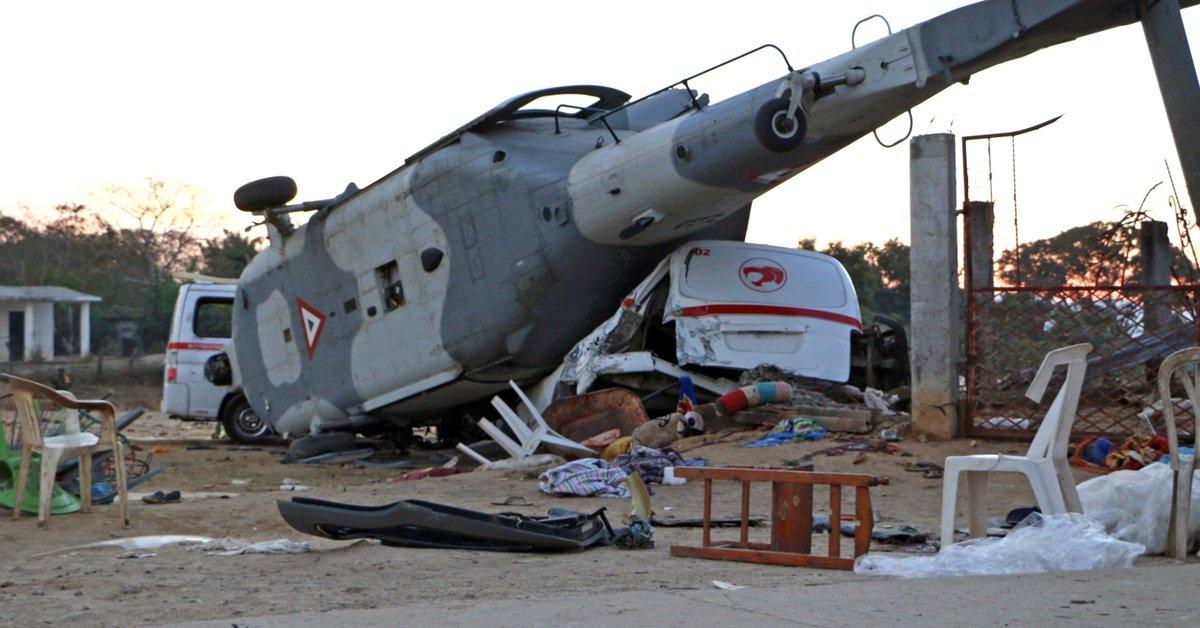}
			\\\\
			{\scriptsize \textbf{Hurricane}}
			&
			{\scriptsize \textbf{Landslide}}
			&
			{\scriptsize \textbf{Other disaster}}
			\\
		\end{tabular}
	}
	
	\caption{Example images for different disaster types. \textit{Not disaster} images are not shown.}
	\label{fig:example_images_disaster_types}
\end{figure*}

When ingesting images from unfiltered social media streams, it is important to automatically detect different disaster types those images show. For instance, an image can depict a wildfire, flood, earthquake, hurricane, and other types of disasters. In the literature, disaster types have been defined in different hierarchical categories such as natural, man-made, and hybrid \cite{shaluf2007disaster}. Natural disasters are events that result from natural phenomena (e.g., fire, flood, earthquake). Man-made disasters are events that result from human actions (e.g., terrorist attack, accidents, war, and conflicts). Hybrid disasters are events that result from human actions, which effect natural phenomena (e.g., deforestation results in soil erosion, and climate change). In this study, we focused on most frequently occurring (see in \cite{shaluf2007disaster}) disaster event types such as {\em (i)} earthquake, {\em (ii)} fire, {\em (iii)} flood, {\em (iv)} hurricane, and {\em (v)} landslide. In addition, we also included two additional class labels such as {\em (vi)} other disaster -- to cover all other disaster types (e.g., plane crash), and {\em (vii)}  not disaster -- for images that do not show any identifiable disasters. This results in a total of seven categories for the disaster type classification task. In Figure~\ref{fig:example_images_disaster_types}, we provide example images for different disaster types. 

\subsection{Informativeness}
\label{ssec:task_informativeness}
Images posted on social media during disasters do not always contain informative (e.g., image showing damaged infrastructure due to flood, fire or any other disaster events) or useful content for humanitarian aid. It is necessary to remove any irrelevant or redundant content to facilitate crisis responders' efforts more effectively. Therefore, the purpose of this classification task is to filter irrelevant images. The class labels for this task are {\em (i)} informative and {\em (ii)} not informative.

\subsection{Humanitarian}
\label{ssec:task_humanitarian}
An important aspect of crisis responders is to assist people based on their needs, which requires information to be classified into more fine-grained categories to take specific actions. In the literature, humanitarian categories often include \textit{affected individuals}; \textit{injured or dead people}; \textit{infrastructure and utility damage}; \textit{missing or found people}; \textit{rescue, volunteering, or donation effort}; and \textit{vehicle damage}~\cite{alam2018crisismmd}. In this study, we focus on four categories that are deemed to be the most prominent and important for crisis responders such as {\em (i)} affected, injured, or dead people, {\em (ii)} infrastructure and utility damage, {\em (iii)} rescue volunteering or donation effort, and {\em (iv)} not humanitarian. 

\subsection{Damage severity}
\label{ssec:task_damage_severity}
Assessing the severity of the damage is important to help the affected community during disaster events. The severity of damage can be assessed based on the physical destruction to a built-structure visible in an image (e.g., destruction of bridges, roads, buildings, burned houses, and forests). Following the work reported in~\cite{nguyen17damage}, we define the categories for this classification task as {\em (i)} severe damage, {\em (ii)} mild damage, and {\em (iii)}~little or none. 

\begin{figure}[t]
\centering
\includegraphics[width=0.30\textwidth]{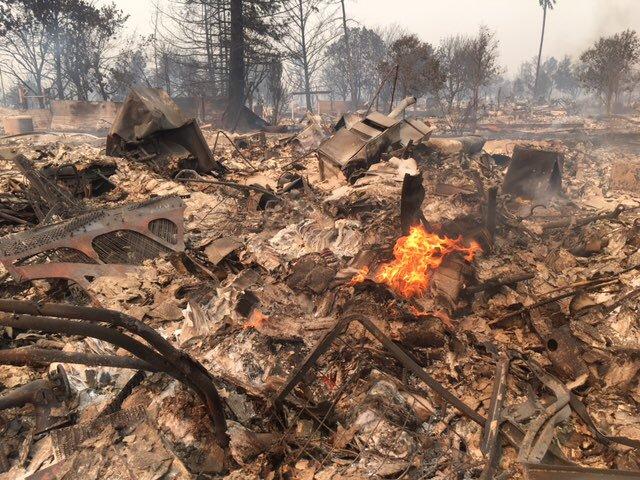}
\vspace{-0.30em}
\caption{An image annotated as {\em (i)}~fire event, {\em (ii)}~informative, {\em (iii)}~infrastructure and utility damage, and {\em (iv)}~severe damage.}
\label{fig:example_all_task}
\vspace{-1.6em}
\end{figure}

Figure \ref{fig:example_all_task} shows an example image that illustrates available annotations for all four tasks. 

\section{Data Preparation}
\label{sec:dataset}

\subsection{Datasets}
\label{ssec:dataset}
For this study, we used public and in-house labeled datasets. Below, we provide the details of each dataset. 

\subsubsection{Damage Assessment Dataset (DAD)}
\label{sssec:dataset_dad}
The damage assessment dataset consists of labeled imagery data with damage severity levels such as severe, mild, and little-to-no damage~\cite{nguyen17damage}. The images have been collected from two sources: AIDR~\cite{imran2014aidr} and Google. To crawl data from Google, authors used the following keywords: damage building, damage bridge, and damage road. The images from AIDR were collected from Twitter during different disaster events such as Typhoon Ruby, Nepal Earthquake, Ecuador Earthquake, and Hurricane Matthew. The dataset contains $\sim25K$ images annotated by paid-workers as well as volunteers. 
In this study, we use this dataset for the informativeness and damage severity tasks. For the informativeness task, we map \textit{mild} and \textit{severe} images into informative class and manually sift through the \textit{little-to-no damage} images to separate them into \textit{informative} and \textit{not informative} categories. For the damage severity task, we map the label \textit{little-to-no damage} into \textit{little or none} to align with other datasets.  

\subsubsection{CrisisMMD}
\label{sssec:dataset_crisismmd}
This is a multimodal (i.e., text and image) and multi-task dataset, which consists of $18,082$ images collected from tweets during seven disaster events crawled by the AIDR system~\cite{alam2018crisismmd}. The data is annotated by crowd-workers using the Figure-Eight platform\footnote{Currently acquired by \url{https://appen.com/}} for three different tasks: {\em (i)} informativeness with binary labels (i.e., informative vs. not informative), {\em (ii)} humanitarian with seven class labels (i.e., infrastructure and utility damage, vehicle damage, rescue, volunteering, or donation effort, injured or dead people, affected individuals, missing or found people, other relevant information and not relevant), {\em (iii)} damage severity assessment with three labels (i.e., severe, mild and little or no damage). 

\subsubsection{AIDR Disaster Type Dataset (AIDR-DT)}
\label{sssec:dataset_aidr_dt}
For disaster type classification task, we annotated images with categories mentioned in Section~\ref{ssec:task_disaster_types}. We obtained tweets from 17 disaster events and 3 general collections, all of which have been collected by the AIDR system. The 17 disaster events include flood, earthquake, fire, hurricane, terrorist-attack, and armed-conflict. The tweets in general collections contains keywords related to natural disasters, human-induced disasters, and security incidents. We crawled images for these collections.
After collecting the images we first remove exact duplicates based on \textit{tweet ids}. Then, we remove exact- and near-duplicates images using a duplicate filtering approach discussed in \cite{alam2019SocialMedia}. From a large number of images of these collections, we sampled $\sim$30K images for annotation. 

The labeling of these images was performed in two steps. First, a set of images were labeled as \textit{earthquake}, \textit{fire}, \textit{flood}, \textit{hurricane}, and \textit{none of these categories}. Then, we selected a sample of $\sim$2200 images, which are labeled as \textit{none of these categories} in the previous step for annotating \textit{not disaster} and \textit{other disaster} categories. The rationale for choosing such a sample number of images was due to limited annotation resources. 


For the landslide category, we crawled images from Google, Bing, and Flickr using keywords landslide, mudslide, ``mud slides'', landslip, ``rock slides'', rockfall, ``land slide'', earthslip, rockslide, and ``land collapse''. As images have been collected from different sources, therefore, it resulted in having duplicates. To take this into account, we applied the same duplicate filtering as before to remove exact- and near-duplicate images. Then, the remaining images were manually labeled as \textit{landslide} and \textit{not landslide}.

For the annotation task, we used the following definitions for the disaster types: 
\begin{enumerate}[(i)]
    \itemsep-0.1em
    \item Earthquake: images showing damaged or destroyed buildings, fractured houses, ground ruptures such as railway lines, roads, airport runways, highways, bridges, and tunnels.
    \item Fire: images showing man-made fires or wildfires (forests, grasslands, brush, and deserts), destroyed forests, houses, or infrastructures. 
    \item Flood: images showing flooded areas, houses, roads, and other infrastructures. 
    \item Hurricane: images showing high winds, a storm surge, heavy rains, collapsed electricity polls, grids, and trees.
    \item Landslide: images showing landslide, mudslide, landslip, rockfall, rockslide, earth slip, and land collapse 
    \item Other disasters: images showing any other disaster types such as plane crash, bus, car, or train accident, explosion, war, and conflicts.
    \item Not disaster: images showing cartoon, advertisement, or anything that cannot be easily linked to any disaster type.
\end{enumerate}

\begin{figure}[t]
\centering
\includegraphics[width=1.0\linewidth]{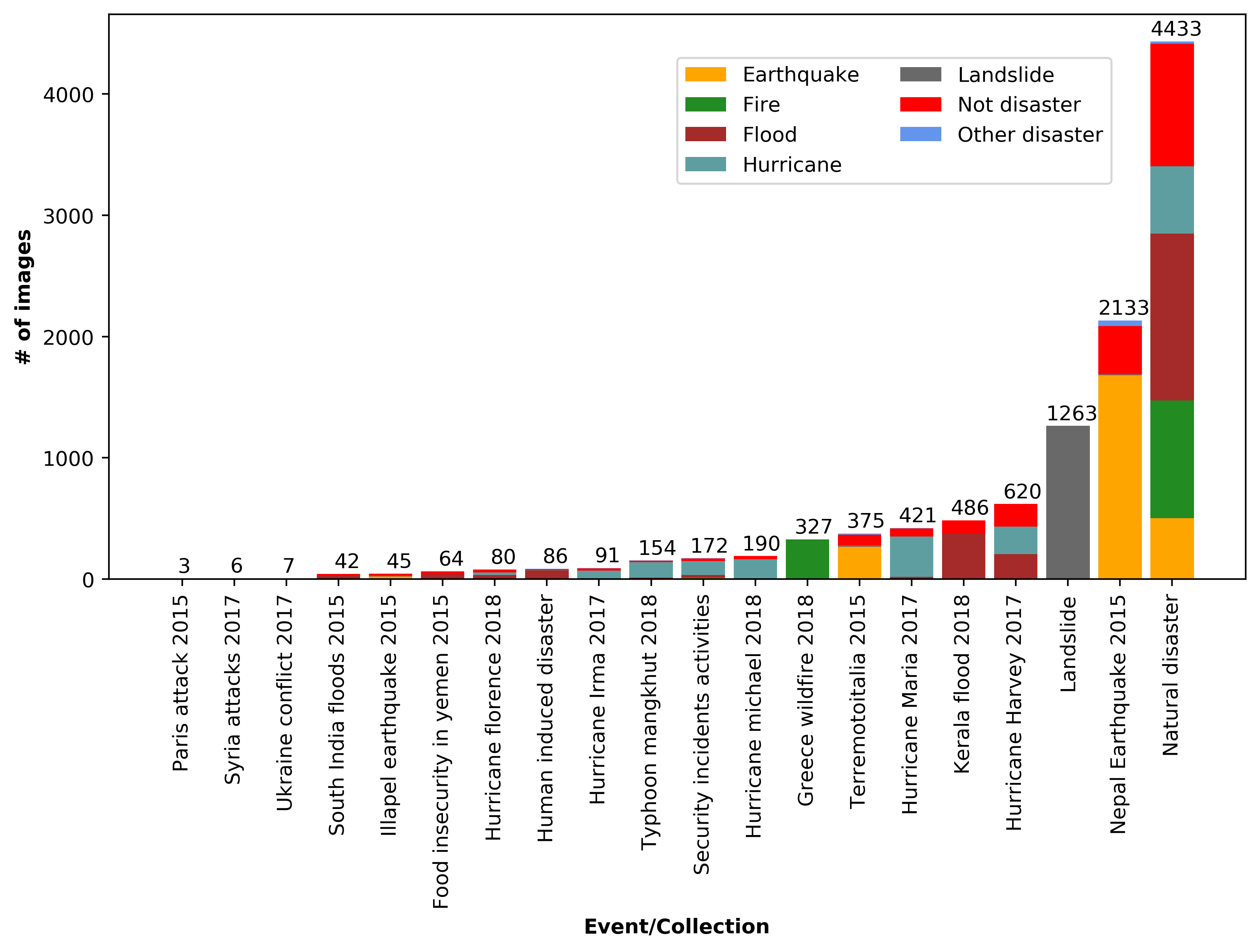}
\caption{Number of disaster types labeled images from different disaster events and collections in AIDR-DT dataset.}
\label{fig:disaster_type_dist}
\end{figure}

In Figure \ref{fig:disaster_type_dist}, we report the distribution of the labeled images in different events and general collections. 


\subsubsection{AIDR Informativeness Dataset (AIDR-Info)}
\label{ssec:dataset_aidr_info}
\begin{figure}[t]
\centering
\includegraphics[width=1.0\linewidth]{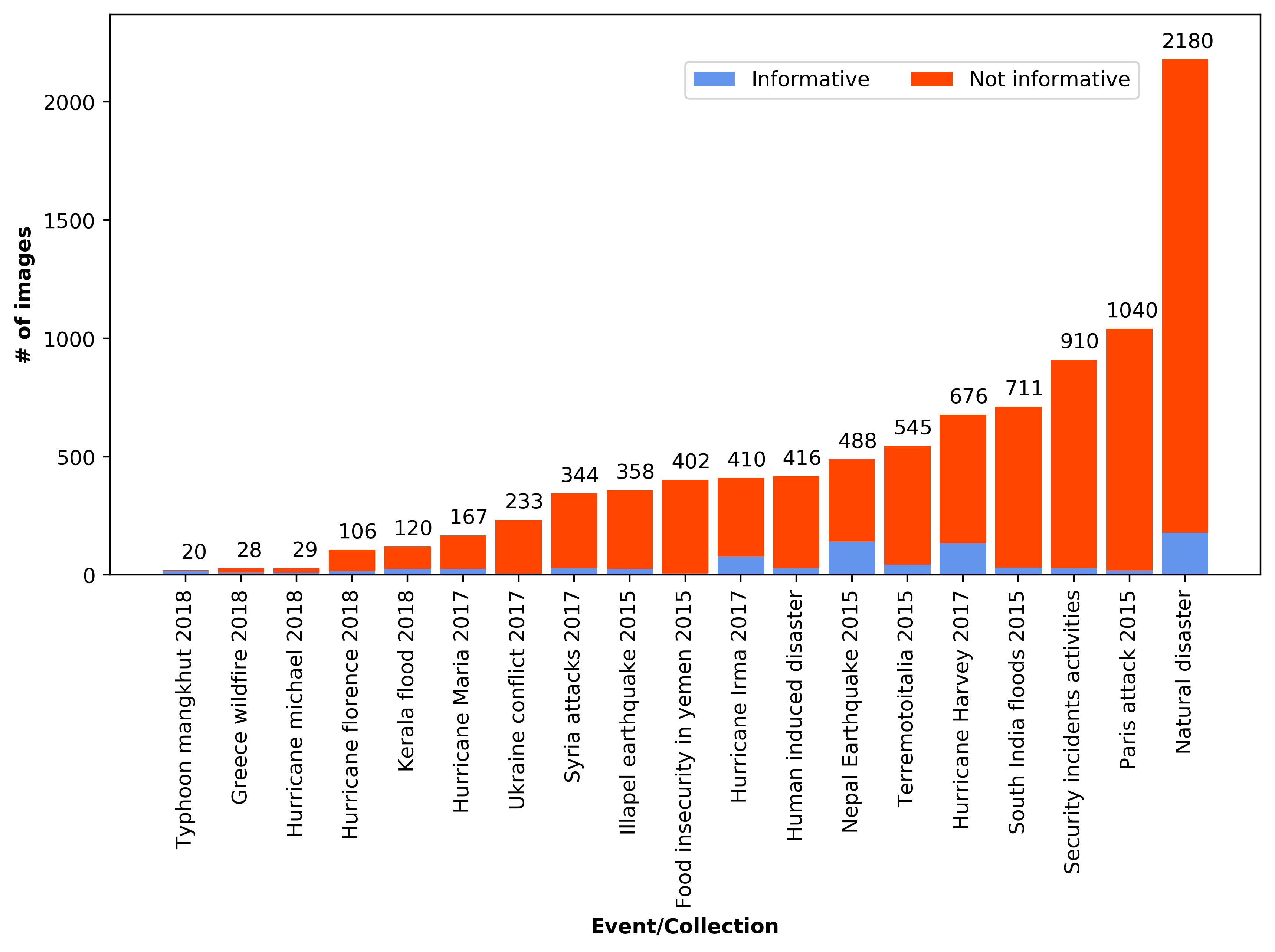}
\caption{Number of labeled informative \textit{vs.} not-informative images from different disaster events and collections in AIDR-info dataset.}
\label{fig:informativeness_dist}
\end{figure}

For this dataset, we collected tweets and images using the AIDR system. We used the same duplicate filtering approach to remove duplicate images. Then, we labeled $9,936$ images with two class labels, informative vs. not-informative using the definition discussed in \cite{alam2018crisismmd}.\footnote{If the image is useful for humanitarian aid then we label it as ``informative'' otherwise as ``not informative''.}  
In Figure \ref{fig:informativeness_dist}, we report the distribution of images labeled for different events. Across different collections, number of not informative images is higher than informative images. 

\subsubsection{Damage Multimodal Dataset (DMD)}
\label{sssec:dataset_dmd}
The multimodal damage identification dataset consists of 5,878
images collected from Instagram and Google~\cite{Mouzannar2018}. Authors of the study crawled the images using more than 100 hashtags, which are proposed in crisis lexicon \cite{olteanu2014crisislex}. The manually labeled data consist of six damage class labels such as fires, floods, natural landscape, infrastructural, human, and non-damage. The non-damage image includes cartoons, advertisements, and images that are not relevant or useful for humanitarian tasks. For this study, we re-labeled them for all four tasks. 
using the same class labels discussed in the previous section. We followed the annotation instructions reported in \cite{alam2018crisismmd} and as discussed in section~\ref{ssec:task_disaster_types}.

\subsection{Annotation}
The annotation has been done by domain experts and ensured the quality of the annotation for new datasets and relabeling existing datasets for new tasks. For the disaster type labeling, the annotators followed the definition discussed in \ref{sssec:dataset_aidr_dt} and for other tasks, definitions and instructions are adapted from \cite{alam2018crisismmd}. 

\subsection{Data Split}
Before consolidating the datasets we split each of them into train, dev, and test sets with 70:10:20 ratio, respectively. The purpose was threefold: {\em (i)}~train and evaluate individual datasets on each task, {\em (ii)}~have a close-to-equal distribution from each dataset into the final consolidated dataset, and {\em (iii)}~provide the research community an opportunity to use the splits independently. After data split, we identify duplicate images (see in Section \ref{ssec:dup_image_identification}) across sets and move them into the training set to create a non-overlapping test set.

\subsection{Duplicate Image Identification}
\label{ssec:dup_image_identification}

To develop a machine learning model, it is important to design non-overlapping training and test sets. A common practice is to randomly split the dataset into train and test sets. This approach often creates an overlapping train-test split with social media data. For example, exact- or near-duplicate images can be in both train and test sets. Based on the work in \cite{alam2019SocialMedia}, we identified duplicate images. Since all datasets have already been manually labeled, we did not want to remove any image from any dataset. We instead attempted to create a non-overlapping train, dev, and test split. The motivation is that having exact- and near-duplicate images in the training set creates a natural augmentation in the training set.

To identify duplicate images in the test set, we first train the model using train and dev set and find the nearest images of the test set. To train the model, we first extract features using a pre-trained deep learning model.\footnote{Note that the pre-trained model is trained using ResNet18 architecture \cite{he2016deep} on the damage assessment dataset~\cite{nguyen17damage}.} Then, we use the Nearest Neighbor \cite{cunningham2007k} to train the model with the training set of each respective dataset. For example, for the informative dataset of CrisisMMD, we use the training set to train the model and then use it to obtain the nearest images for each image in the test set. 

\begin{figure}[t]
\centering
\includegraphics[width=0.9\linewidth]{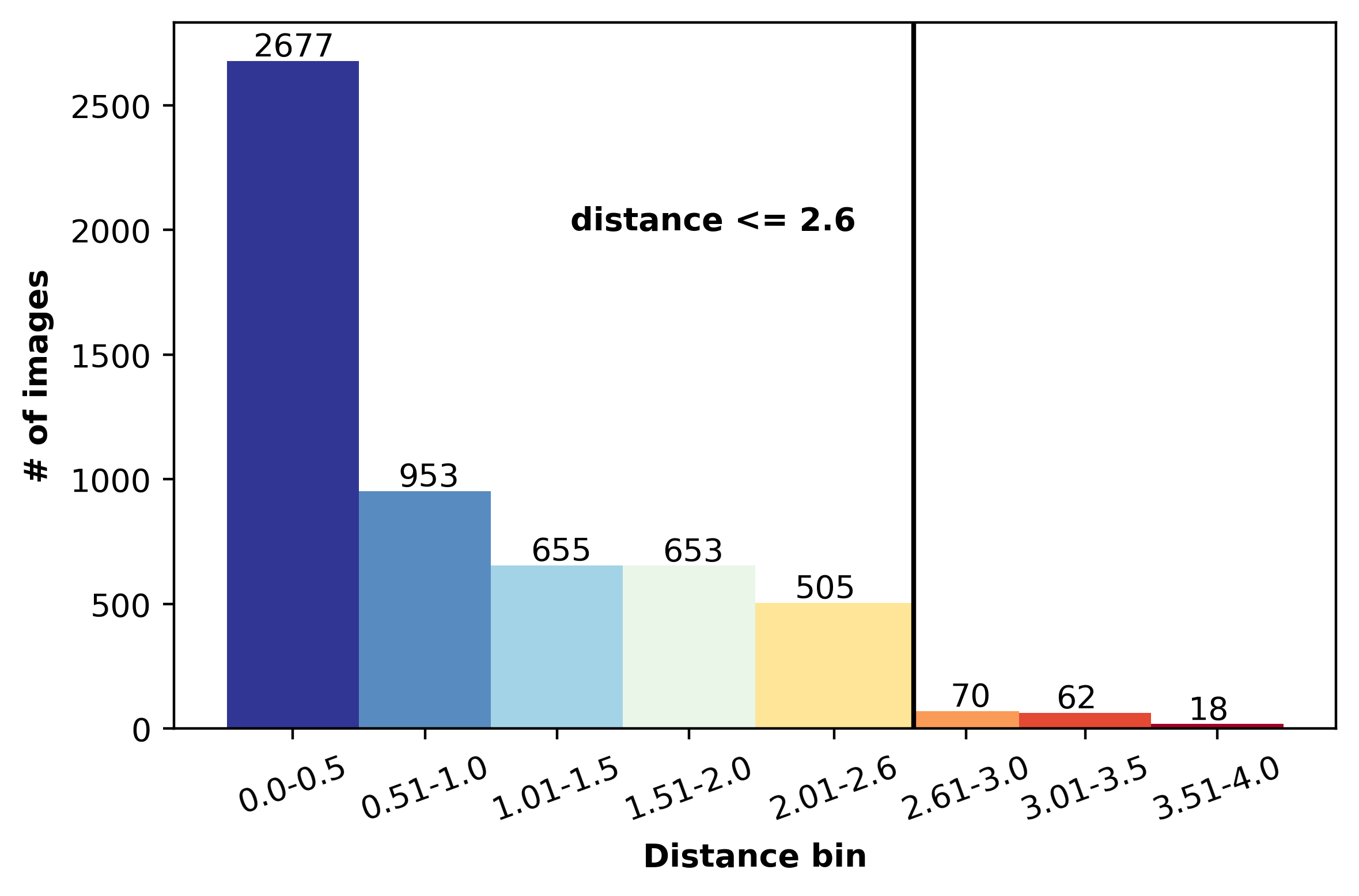}
\caption{Number of duplicates in different distance bins.}
\label{fig:duplicates_with_bins}
\end{figure}

Next, we manually identify duplicates by investigating each image from a given test set and the identified nearest images from the corresponding train set 
for four different tasks and twelve different datasets. 
Out of these images, we identified 5,593 exact- and near-duplicate images in different test sets. We then move the identified images to the training set to create non-overlapping test sets. It also helped us to identify an approximate threshold to automatically identify near-duplicate images. In Figure~\ref{fig:duplicates_with_bins}, we present a histogram of \textit{Euclidean} distance measures of the exact- and near-duplicate images. It shows the number of images in different bins. 
With our analysis, we realized that a 
distance threshold of less than equal $2.6$ is a reasonable choice for automatic duplicate detection. From the figure, we observe that there are also duplicate images with higher thresholds; however, that is a very small number comparatively. Also, note that choosing a higher number will lead to an increase in false positives. Using the threshold value of $d\leq2.6$ we automatically identified duplicate images in the dev set 
and moved them to the train set. 


It is important to note that creating non-overlapping datasets using duplicate identification process reduces the distribution of dev and test sets. This is reasonable given the fact that we are ensuring unbiased train/test splits.


\subsection{Data Consolidation:}
\label{ssec:data_consolidation}
One of the important reasons to perform data consolidation is to develop robust deep learning models with large amounts of data. For this purpose, we merge all train, dev, and test sets into the consolidated train, dev, and test sets, respectively. As combining multiple datasets can results in duplicate images in train and test set, after merging the dataset, we repeat the same duplicate identification procedure to create non-overlapping sets for different tasks.


\subsection{Data Statistics}
\label{ssec:class_label_dist}
Tables \ref{tab:data_split_disaster_types}, \ref{tab:data_split_informativemess}, \ref{tab:data_split_humanitarian}, \ref{tab:data_split_damage_severity}, and \ref{tab:consolidated_dataset} show the label distribution of all datasets for different tasks. We report the total number of images in parenthesis for each dataset in the Tables. Some class labels are skewed in individual datasets. For example, in disaster type datasets (Table \ref{tab:data_split_disaster_types}), the distribution of ``other disaster'' label is low in AIDR-DT dataset, whereas the distribution of ``landslide'' label low in DMD dataset. For the informativeness task, low distribution is observed for the ``informative" label. Moreover, for the humanitarian task, we have low distribution for ``rescue volunteering or donation effort'' label in DMD dataset, and for the damage severity task ``mild" label in CrisisMMD and DMD datasets. However, the consolidated dataset creates a fair balance across class labels for different tasks as shown in Table~\ref{tab:consolidated_dataset}.

\begin{table}[t]
\centering
\caption{Data split for the \textbf{disaster types} task. Number in parenthesis shows total number of images.}
\vspace{-0.5em}
\label{tab:data_split_disaster_types}
\scalebox{0.85}{
\begin{tabular}{@{}llrrrr@{}}
\toprule
\multicolumn{1}{@{}l}{\textbf{Dataset}} & \multicolumn{1}{l}{\textbf{Class labels}} & \multicolumn{1}{r}{\textbf{Train}} & \multicolumn{1}{r}{\textbf{Dev}} & \multicolumn{1}{r}{\textbf{Test}} & \multicolumn{1}{r@{}}{\textbf{Total}} \\ \midrule
\multirow{8}{*}{\textbf{\begin{tabular}[c]{@{}l@{}}AIDR-DT \\(11723)\end{tabular}}} & Earthquake & 1910 & 201 & 376 & 2487 \\
 & Fire & 990 & 105 & 214 & 1309 \\
 & Flood & 2059 & 241 & 533 & 2833 \\
 & Hurricane & 1188 & 142 & 279 & 1609 \\
 & Landslide & 901 & 119 & 257 & 1277 \\
 & Not disaster & 1507 & 198 & 415 & 2120 \\
 & Other disaster & 65 & 6 & 17 & 88 \\\midrule 
\multirow{8}{*}{\textbf{\begin{tabular}[c]{@{}l@{}}DMD \\(5788)\end{tabular}}} & Earthquake & 130 & 17 & 35 & 182 \\
 & Fire & 255 & 36 & 71 & 362 \\
 & Flood & 263 & 35 & 70 & 368 \\
 & Hurricane & 253 & 36 & 73 & 362 \\
 & Landslide & 38 & 5 & 11 & 54 \\
 & Not disaster & 2108 & 288 & 575 & 2971 \\
 & Other disaster & 1057 & 145 & 287 & 1489 \\ 
 \bottomrule 
\end{tabular}
}
\vspace{-0.8em}
\end{table}

\begin{table}[t]
\centering
\caption{Data split for the \textbf{informativeness} task.}
\vspace{-0.5em}
\label{tab:data_split_informativemess}
\scalebox{0.85}{
\begin{tabular}{@{}llrrrr@{}}
\toprule
\multicolumn{1}{@{}l}{\textbf{Dataset}} & \multicolumn{1}{l}{\textbf{Class labels}} & \multicolumn{1}{r}{\textbf{Train}} & \multicolumn{1}{r}{\textbf{Dev}} & \multicolumn{1}{r}{\textbf{Test}} & \multicolumn{1}{r@{}}{\textbf{Total}} \\ \midrule
\multirow{2}{*}{\textbf{\begin{tabular}[c]{@{}l@{}}DAD \\(25820)\end{tabular}}} & Informative & 15329 & 590 & 2266 & 18185 \\
 & Not informative & 5950 & 426 & 1259 & 7635 \\ \midrule 
\multirow{2}{*}{\textbf{\begin{tabular}[c]{@{}l@{}}CrisisMMD \\(18082)\end{tabular}}} & Informative & 7233 & 635 & 1507 & 9375 \\
 & Not informative & 6535 & 551 & 1621 & 8707 \\ \midrule 
\multirow{2}{*}{\textbf{\begin{tabular}[c]{@{}l@{}}DMD \\(5878)\end{tabular}}} & Informative & 2071 & 262 & 573 & 2906 \\
 & Not informative & 2152 & 240 & 580 & 2972 \\ \midrule 
\multirow{2}{*}{\textbf{\begin{tabular}[c]{@{}l@{}}AIDR-Info \\(9936)\end{tabular}}} & Informative & 627 & 66 & 172 & 865 \\
 & Not informative & 6677 & 598 & 1796 & 9071 \\ 
 \bottomrule
 \end{tabular}
}
\vspace{-0.8em}
\end{table}

\begin{table}[t]
\centering
\caption{Data split for the \textbf{humanitarian} task.}
\vspace{-0.5em}
\label{tab:data_split_humanitarian}
\scalebox{0.8}{
\begin{tabular}{@{}llrrrr@{}}
\toprule
\multicolumn{1}{@{}l}{\textbf{Dataset}} & \multicolumn{1}{l}{\textbf{Class labels}} & \multicolumn{1}{r}{\textbf{Train}} & \multicolumn{1}{r}{\textbf{Dev}} & \multicolumn{1}{r}{\textbf{Test}} & \multicolumn{1}{r@{}}{\textbf{Total}} \\ \midrule
\multirow{5}{*}{\textbf{\begin{tabular}[c]{@{}l@{}}CrisisMMD\\ (11241)\end{tabular}}} & Affected, injured, or dead people & 521 & 51 & 100 & 672 \\
 & Infrastructure and utility damage & 3040 & 299 & 589 & 3928 \\
 & Not humanitarian & 3307 & 296 & 807 & 4410 \\
 & Rescue volunteering or donation effort & 1682 & 174 & 375 & 2231 \\ \midrule 
\multirow{5}{*}{\textbf{\begin{tabular}[c]{@{}l@{}}DMD\\ (5528)\end{tabular}}} & Affected, injured, or dead people & 242 & 28 & 63 & 333 \\
 & Infrastructure and utility damage & 933 & 125 & 242 & 1300 \\
 & Not humanitarian & 2736 & 314 & 744 & 3794 \\
 & Rescue volunteering or donation effort & 74 & 9 & 18 & 101 \\ 
 \bottomrule
\end{tabular}
}
\vspace{-0.5em}
\end{table}

\begin{table}[t]
\centering
\caption{Data split for the \textbf{damage severity} task.}
\label{tab:data_split_damage_severity}
\scalebox{0.85}{
\begin{tabular}{@{}llrrrr@{}}
\toprule
\multicolumn{1}{@{}l}{\textbf{Dataset}} & \multicolumn{1}{l}{\textbf{Class labels}} & \multicolumn{1}{r}{\textbf{Train}} & \multicolumn{1}{r}{\textbf{Dev}} & \multicolumn{1}{r}{\textbf{Test}} & \multicolumn{1}{r@{}}{\textbf{Total}} \\ \midrule
\multirow{4}{*}{\textbf{\begin{tabular}[c]{@{}l@{}}DAD\\ (25820)\end{tabular}}} & Little or none & 7881 & 1101 & 1566 & 10548 \\
 & Mild & 2828 & 388 & 546 & 3762 \\
 & Severe & 9457 & 673 & 1380 & 11510 \\ \midrule 
\multirow{4}{*}{\textbf{\begin{tabular}[c]{@{}l@{}}CrisisMMD\\ (3198)\end{tabular}}} & Little or none & 317 & 35 & 67 & 419 \\
 & Mild & 547 & 56 & 125 & 728 \\
 & Severe & 1629 & 144 & 278 & 2051 \\\midrule 
\multirow{4}{*}{\textbf{\begin{tabular}[c]{@{}l@{}}DMD \\(5878)\end{tabular}}} & Little or none & 2874 & 331 & 778 & 3983 \\
 & Mild & 508 & 60 & 132 & 700 \\
 & Severe & 857 & 110 & 228 & 1195 \\ 
 \bottomrule
\end{tabular}
}
\vspace{-0.5em}
\end{table}

\begin{table}[t]
\centering
\caption{Data splits for the \textbf{consolidated dataset} for all tasks.}
\vspace{-0.5em}
\label{tab:consolidated_dataset}
\scalebox{0.85}{
\begin{tabular}{@{}lrrrr@{}}
\toprule
\multicolumn{1}{@{}l}{\textbf{Class labels}} & \multicolumn{1}{r}{\textbf{Train}} & \multicolumn{1}{r}{\textbf{Dev}} & \multicolumn{1}{r}{\textbf{Test}} & \multicolumn{1}{r@{}}{\textbf{Total}} \\ \midrule
\multicolumn{5}{c}{\textbf{Disaster types (17511)}} \\ \midrule
Earthquake & 2058 & 207 & 404 & 2669 \\
Fire & 1270 & 121 & 280 & 1671 \\
Flood & 2336 & 266 & 599 & 3201 \\
Hurricane & 1444 & 175 & 352 & 1971 \\
Landslide & 940 & 123 & 268 & 1331 \\
Not disaster & 3666 & 435 & 990 & 5091 \\
Other disaster & 1132 & 143 & 302 & 1577 \\ \midrule 
\multicolumn{5}{c}{\textbf{Informativeness (59717)}} \\\midrule
Informative & 26486 & 1432 & 3414 & 31332 \\
Not informative & 21700 & 1622 & 5063 & 28385 \\ \midrule 
\multicolumn{5}{c}{\textbf{Humanitarian (16769)}} \\ \midrule 
Affected, injured, or dead people & 772 & 73 & 160 & 1005 \\
Infrastructure and utility damage & 4001 & 406 & 821 & 5228 \\
Not humanitarian & 6076 & 578 & 1550 & 8204 \\
Rescue volunteering or donation effort & 1769 & 172 & 391 & 2332 \\ \midrule  
\multicolumn{5}{c}{\textbf{Damage severity (34896)}} \\\midrule
Little or none & 11437 & 1378 & 2135 & 14950 \\
Mild & 4072 & 489 & 629 & 5190 \\
Severe & 12810 & 845 & 1101 & 14756 \\ 
\bottomrule
\end{tabular}
}
\vspace{-0.7em}
\end{table}
\section{Experiments}
\label{sec:experiments}

\subsection{Experimental Settings}
We employ the transfer learning approach to perform experiments, which has shown promising results for various visual recognition tasks in the literature~\cite{yosinski2014transferable,sharif2014cnn,ozbulak2016transferable,oquab2014learning}. The idea of the transfer learning approach is to use existing weights of a pre-trained model. For this study, we used several neural network architectures using the PyTorch library.\footnote{https://pytorch.org/} The architectures include ResNet18, ResNet50, ResNet101~\cite{he2016deep}, AlexNet~\cite{krizhevsky2012imagenet}, VGG16~\cite{simonyan2014very},  DenseNet~\cite{huang2017densely}, SqueezeNet~\cite{i2016squeezenet}, InceptionNet~\cite{szegedy2016rethinking}, MobileNet~\cite{howard2017mobilenets}, and EfficientNet~\cite{tan2019efficientnet}.  

We use the weights of the networks trained using ImageNet~\cite{deng2009imagenet} to initialize our model. 
We adapt the last layer (i.e., softmax layer) of the network according to the particular classification task at hand instead of the original 1,000-way classification. The transfer learning approach allows us to transfer the features and the parameters of the network from the broad domain (i.e., large-scale image classification) to the specific one, in our case four different classification tasks.
We train the models using the Adam optimizer~\cite{kingma2014adam} 
with an initial learning rate of $10^{-5}$, which is decreased by a factor of 10 
when accuracy on the dev set stops improving for 10 epochs. 

We designed the binary classifier for informativeness task and multiclass classifiers for other tasks.

To measure the performance of each classifier, we use weighted average precision (P), recall (R), and F1-measure (F1). We only report F1-measure due to limited space. 


\subsection{Datasets Comparison}
To determine whether consolidated data helps achieve better performance, we train the models using training sets from the individual and consolidated datasets. However, we always test the models on the consolidated test set. As our test data is same across different experiments, results are ensured to be comparable. Since we have four different tasks, which consist of fifteen different datasets, we only experimented with the ResNet18~\cite{he2016deep} network architecture to manage the computational load.

\subsection{Network Architectures}
Currently available neural network architectures come with different computational complexity. As one of our goals is to deploy the models in real-time applications, we exploit them to understand their performance differences. Another motivation is that current literature in crisis informatics only reports results using one or two network architectures (e.g., VGG16 in \cite{multimodalbaseline2020}, InceptionNet in \cite{Mouzannar2018}), which we wanted to extend in this study. 

\section{Results and Discussions}
\label{sec:results_discussion_future_works}

\subsection{Results}
\label{ssec:baseline_results}

\subsubsection{Dataset Comparison}
\label{sssec:baseline_results_datasets}
In Table~\ref{tab:classification_results}, we report classification results for different tasks and different datasets using ResNet18 network architecture. The performance of different tasks is not equally comparable as they have different levels of complexity (e.g., varying number of class labels, class imbalance, etc.). For example, the informativeness classification is a binary task, which is computationally simpler than a classification task with more labels (e.g., seven labels in disaster types). Hence, the performance is comparatively higher for informativeness. An example of a class imbalance issue can be seen in Table \ref{tab:consolidated_dataset} with the damage severity task. The distribution of mild is comparatively small, which reflects on its and overall performance. The mild class label is also less distinctive than other class labels, and we noticed that classifiers often confuse this class label with the other two class labels. Similar findings have also been reported in \cite{nguyen17damage}.  
For the disaster types task, the performance of the AIDR-DT model is higher compared to the DMD model. We observe that the DMD dataset is comparatively small and the model is not performing well on the consolidated dataset. This characteristic is observed in other tasks as well. As expected, overall for all tasks, the models with the consolidated datasets outperform individual datasets.


\begin{table}[]
\centering
\caption{Results of different classification tasks using the ResNet18 models. Trained on individual and consolidated datasets and tested on consolidated test sets. Disaster types (Disas.), Informativeness (Info.),  Humanitarian (Hum.), Damage severity (Damage).}
\vspace{-0.5em}
\label{tab:classification_results}
\scalebox{0.9}{
\begin{tabular}{@{}lrrrr@{}}
\toprule
\multicolumn{1}{@{}l}{\textbf{Dataset}} & \multicolumn{1}{r}{\textbf{Disas.}} & \multicolumn{1}{r}{\textbf{Info.}} & \multicolumn{1}{r}{\textbf{Hum.}} & \multicolumn{1}{r@{}}{\textbf{Damage}} \\ \midrule
AIDR-DT & 0.726 & - & - & - \\
DAD & - & 0.797 & - & 0.709 \\
CrisisMMD & - & 0.790 & 0.727 & 0.374 \\
DMD & 0.591 & 0.799 & 0.636 & 0.663 \\
AIDR-Info & - & 0.725 & - & - \\
Consolidated & \underline{\textbf{0.785}} & \underline{\textbf{0.851}} & \underline{\textbf{0.749}} & \underline{\textbf{0.736}} \\ \bottomrule
\end{tabular}
}
\vspace{-1.5em}
\end{table}

\subsubsection{Network Architectures Comparison}
\label{sssec:results_network_comparison}
In Table~\ref{tab:classification_results_net_comparison}, we report results using different network architectures on consolidated datasets for different tasks, i.e., trained and tested using a consolidated dataset. Across different tasks, overall EfficientNet is performing better than other models except for humanitarian task, for which VGG16 is outperforming other models. Comparatively the second-best models are VGG16, ResNet50, ResNet101, and DenseNet (101). From the results of different tasks, we observe that InceptionNet is the worst performing model. 

In Table~\ref{tab:classification_results_net_comparison}, we also report different neural network models with their number of layers, parameters, and memory consumption during inference. There can always be a trade-off between performance vs.\ computational complexity, i.e., number of layers, parameters, and memory consumption. In terms of memory consumption and the number of parameters, VGG16 seems expensive than others. Based on the performance and computational complexity, we can conclude that EfficientNet can be the best option to use in real-time applications. We computed throughput for EfficientNet using a batch size of 128 and it can process ${\sim}$260 images per second on an NVIDIA Tesla P100 GPU. 
Among different ResNet models, ResNet18 is a reasonable choice given that its computational complexity is significantly less than other ResNet models.

\begin{table}[]
\centering
\caption{Results using different neural network models on the consolidated dataset with four different tasks. Trained and tested using the consolidated dataset. Comparable result is shown in \textbf{bold} and best results is shown in \underline{underlined}. \#L (P) - number of layers (number of parameters in millions). Mem. (memory in MB). IncepNet (InceptionNet), MobNet (MobileNet), EffiNet (EfficientNet)}
\vspace{-0.5em}
\label{tab:classification_results_net_comparison}
\scalebox{0.88}{
\begin{tabular}{@{}lrrrrrrr@{}}
\toprule
\multicolumn{1}{@{}}{\textbf{Model}} & \multicolumn{1}{c}{\textbf{\# L (P)}} & \multicolumn{1}{r}{\textbf{Mem.}} & \multicolumn{1}{r}{\textbf{Disas.}} & \multicolumn{1}{r}{\textbf{Info.}} & \multicolumn{1}{r}{\textbf{Hum.}} & \multicolumn{1}{r}{\textbf{Damage}} & \multicolumn{1}{r@{}}{\textbf{Avg.}} \\ \midrule
ResNet18 & 18 (11.2) & 74.6 & 0.785 & 0.851 & 0.749 & 0.736 & 0.780 \\
ResNet50 & 50 (23.5) & 233.5 & \textbf{0.808} & 0.852 & 0.762 & \textbf{0.751} & \textbf{0.793} \\
ResNet101 & 101 (42.5) & 377.6 & \textbf{0.813} & 0.852 & \textbf{0.765} & 0.737 & \textbf{0.792} \\
AlexNet & 8 (57.0) & 222.2 & 0.754 & 0.828 & 0.716 & 0.709 & 0.752 \\
VGG16 & 16 (134.3) & 673.9 & 0.798 & \textbf{0.858} & \textbf{0.773} & \textbf{0.753} & \textbf{0.796} \\
DenseNet & 121 (7.0) & 174.2 & \textbf{0.806} & \textbf{0.862} & 0.755 & 0.739 & \textbf{0.791} \\
SqueezeNet & 18 (0.7) & 48.0 & 0.755 & 0.829 & 0.719 & 0.708 & 0.753 \\
IncepNet & 42 (24.3) & 206.0 & 0.528 & 0.593 & 0.509 & 0.615 & 0.561 \\
MobNet (v2) & 20 (2.2) & 8.5 & 0.782 & 0.849 & 0.746 & 0.730 & 0.777 \\
EffiNet (b1) & 25 (7.8) & 177.8 & \underline{\textbf{0.816}} & \underline{\textbf{0.863}} & \underline{\textbf{0.765}} & \underline{\textbf{0.758}} & \underline{\textbf{0.801}} \\ \bottomrule
\end{tabular}
}
\vspace{-0.7em}
\end{table}


\subsection{Discussions}
\label{ssec:discussions}
Achieving a better performance with deep learning models requires relatively larger datasets. To date, the developed dataset sizes for disaster response tasks are comparatively small. Hence, we address that by combining data from multiple sources and relabeled them for new tasks. The proposed datasets consists of binary and multiple class labels and addressed in binary and multiclass classification settings. However, the datasets can be turned into multi-label and multi-task settings, which we aim to address in a future study.

A significant challenge with social media data is the exact- and near-duplicate content. 
We address this issue, and our proposal for the community is to remove duplicates before the annotation process. Towards this direction, another important challenge is that current duplicate detection is similarity and threshold-based with deep learning feature extraction. In our analysis, we describe a procedure to determine a reasonable threshold for automatic duplicate detection. However, this requires further study, which we aim to do in the future.

Real-time event detection is an important problem from social media content. Our new disaster types dataset can help to develop models and deploy in real-time applications. We also explore several deep learning models, which vary with performance and complexities. Among them, EfficientNet appears to be a reasonable option. Note that EfficientNet has a series of network architectures (b0-b7) and for this study, we only reported results with EfficientNet (b1). We aim to further explore other architectures. 

A small and low latency model is desired to deploy mobile and handheld embedded computer vision applications. The development of MobileNet~\cite{howard2017mobilenets} sheds light towards that direction. Our experimental results suggest that it is computationally simpler and provides a reasonable accuracy, only 2-3\% lower than the best models for different tasks.

Comparing our results with previous state-of-the-art is not possible due to differences in data splits and the issue of duplicate images. On informativeness and humanitarian tasks, previous reported results (weighted F1) are 83.2 and 76.3, respectively, using the CrisisMMD dataset \cite{multimodalbaseline2020}. The authors in \cite{Mouzannar2018} reported a test accuracy of $83.98\pm1.72$ for six disaster types tasks using the DMD dataset with a five-fold cross-validation run. In another study, using the CrisisMMD dataset, authors report weighted-F1 of 81.22 and 86.96 for informativeness and humanitarian tasks, respectively~\cite{abavisani2020multimodal}. They used a small subset of the whole CrisisMMD dataset in their study. Due to differences in data splits, these systems are hard to compare. However, we hope our datasets and splits will provide a standard ground for future studies to compare results. 
 

\section{Conclusions}
\label{sec:conclutions}
Images shared on social media contain useful information for humanitarian organizations. There has been limited work for disaster response image classification tasks compared to text due to the limited resources to develop deep learning models. In this study, we provide new datasets for disaster type detection and informativeness classification. We also relabeled existing datasets for new tasks, and provide a consolidated dataset. We identified duplicates and created non-overlapping splits, which can ensure unbiased results. We addressed four tasks such as disaster types, informativeness, humanitarian and damage severity, that are needed for disaster response. The datasets have a unique characteristic that it can turn into multi-label, and multitask learning setups and would be useful for the deep learning community to develop new algorithms. We also aim to address this in the future. Furthermore, we used different state-of-the-art deep learning architectures to provide benchmark results on the datasets.
\vspace{-0.1cm}

\bibliographystyle{./bibliography/IEEEtran}
\bibliography{./bibliography/main}


\end{document}